\documentclass{article}

\PassOptionsToPackage{numbers}{natbib}
\usepackage[final]{vln_arxiv}

\usepackage{pdfpages}

\usepackage[utf8]{inputenc} % allow utf-8 input
\usepackage[T1]{fontenc}    % use 8-bit T1 fonts
\usepackage{hyperref}       % hyperlinks
\usepackage{url}            % simple URL typesetting
\usepackage{booktabs}       % professional-quality tables
\usepackage{amsfonts}       % blackboard math symbols
\usepackage{nicefrac}       % compact symbols for 1/2, etc.
\usepackage{microtype}      % microtypography

\usepackage{amsmath}

\usepackage{color}
\usepackage{floatrow}
\usepackage{appendix}
\newcommand{\mytilde}{\raise.17ex\hbox{$\scriptstyle\mathtt{\sim}$}}

\title{Video Ladder Networks}

\author{
  Francesco Cricri\thanks{Equal contribution}\\
  Nokia Technologies\\
  Tampere, Finland \\
  \texttt{francesco.cricri@nokia.com} \\
\And
  Xingyang Ni\footnotemark[1]\\
  Tampere University of Technology\\
  Tampere, Finland \\
  \texttt{xingyang.ni@tut.fi}\\
\And
  Mikko Honkala\\
  Nokia Technologies\\
  Espoo, Finland \\
  \texttt{mikko.honkala@nokia.com} \\
\And
  Emre Aksu\\
  Nokia Technologies\\
  Tampere, Finland \\
  \texttt{emre.aksu@nokia.com} \\
\And
  Moncef Gabbouj\\
  Tampere University of Technology\\
  Tampere, Finland \\
  \texttt{moncef.gabbouj@tut.fi}\\
}

\begin{document}
% \xxxfinalcopy is no longer used

\maketitle

\begin{abstract}
We present the Video Ladder Network (VLN) for efficiently generating future video frames.
VLN is a neural encoder-decoder model augmented at all layers by both recurrent and feedforward lateral connections. At each layer, these connections form a lateral recurrent residual block, where the feedforward connection represents a skip connection and the recurrent connection represents the residual. Thanks to the recurrent connections, the decoder can exploit temporal summaries generated from all layers of the encoder. This way, the top layer is relieved from the pressure of modeling lower-level spatial and temporal details. Furthermore, we extend the basic version of VLN to incorporate ResNet-style residual blocks in the encoder and decoder, which help improving the prediction results. VLN is trained in self-supervised regime on the Moving MNIST dataset, achieving competitive results while having very simple structure and providing fast inference.
\end{abstract}

\section{Introduction}
\label{sec:intro}
In recent years, several research groups in the deep learning and computer vision communities have targeted video prediction (\citet{srivastava2015unsupervised}, \citet{Brabandere2016nips}, \citet{kalchbrenner2016video}). The task consists of providing a model with a sequence of past frames, and asking it to generate the next frames in the sequence (also referred to as the \textit{future} frames). This is a very challenging task, as the model needs to embed a rich internal representation of the world and its physical rules, which are highly-structured. Machine learning -based models for video prediction are typically trained in self-supervised regime, where the ground-truth future frames are provided as targets. This is sometimes referred to also as unsupervised training, as no manually-labelled data are needed.

Training a model to predict future video frames is beneficial for a number of applications. First of all, the learned internal representations can be used for extracting rich semantic features in both space and time, which can then be utilized for different supervised discriminative tasks such as action or activity recognition (as in \citet{srivastava2015unsupervised}), semantic segmentation, etc. Also, the ability to predict (or imagine) the future is an important skill of humans, that it used for anticipating the consequence of actions in the real-world, thus allowing us to make decisions about which action to perform. Hence, these internal video representations may support robots in their action decision process. 

We propose the Video Ladder Network (VLN), a neural network architecture for generating future video frames efficiently. The proposed network is evaluated on a benchmark dataset, the Moving MNIST dataset, and is compared to most relevant prior works. We show that our model achieves competitive results, while having simple structure and providing fast inference.

The rest of the paper is organized as follows. Section \ref{sec:proposed} describes the proposed model. Section \ref{sec:experiments} provides experimental results. Finally, Section \ref{sec:conclusions} draws some concluding remarks.

\section{Proposed Model}
\label{sec:proposed}
\begin{figure}[htbp]
  \centering
  {\includegraphics[width=0.7\linewidth]{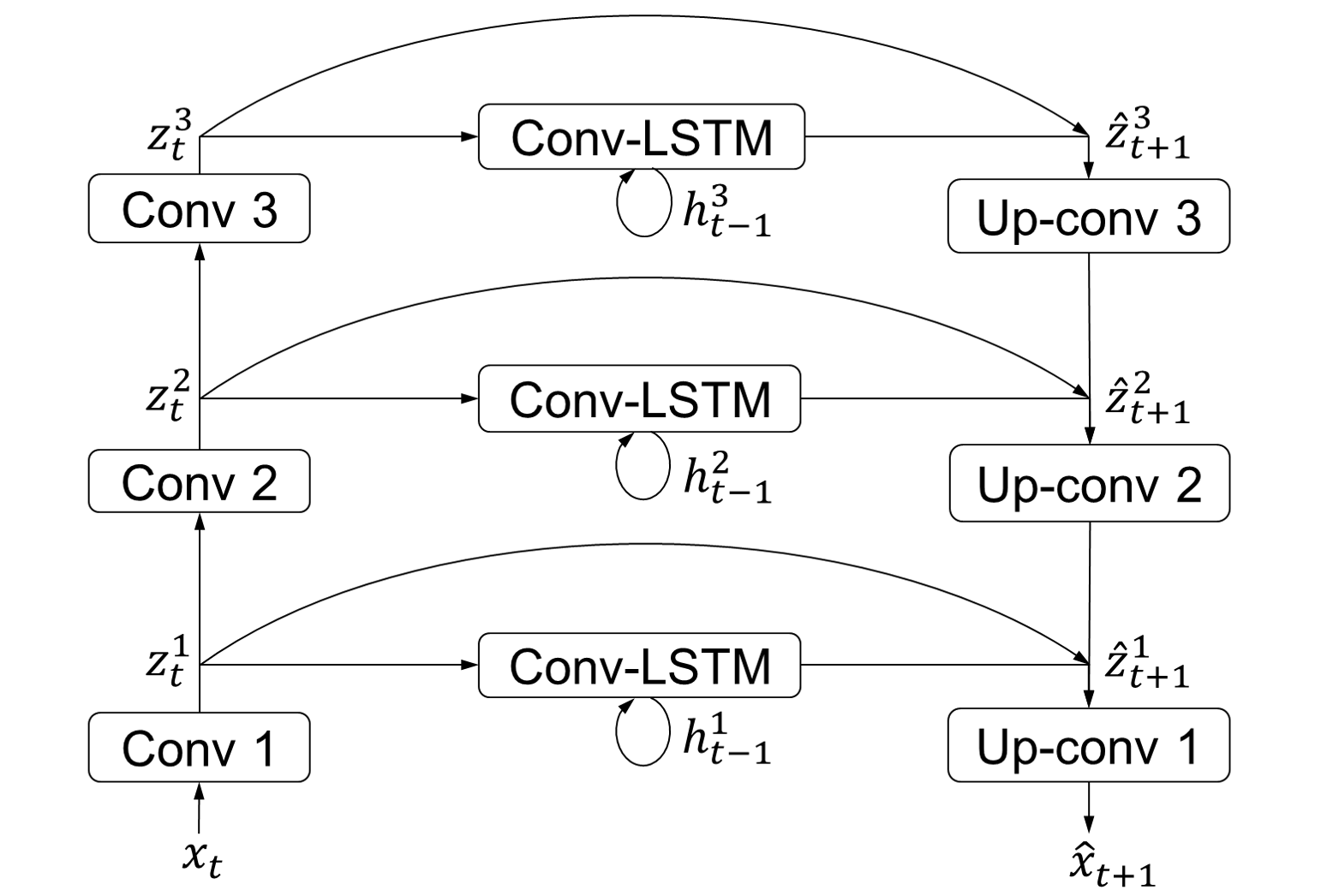}} 
  {\caption{Overview of VLN. $x_t$ is a frame at time $t$. $\hat{x}_{t+1}$ is a predicted frame at time $t+1$}\label{fig:overview}}
\end{figure}

The Video Ladder Network is a fully convolutional neural encoder-decoder network augmented at all layers by specifically-designed \textit{lateral connections} (see Figure \ref{fig:overview}). 

\subsection{Encoder and Decoder}
\label{sec:encdec}
The encoder and decoder of VLN are common fully-convolutional feedforward neural networks. In particular, the encoder consists of dilated convolutional layers (\citet{Yu2015}), whereas the decoder uses normal convolutions. Both encoder and decoder use batch-normalization (BN) (\citet{Ioffe2015}) and \textit{leaky-ReLU} activation function (except for the last decoder's layer, which uses a \textit{sigmoid} activation).

\begin{figure}[htbp]
  \centering
  {\includegraphics[width=1.0\linewidth]{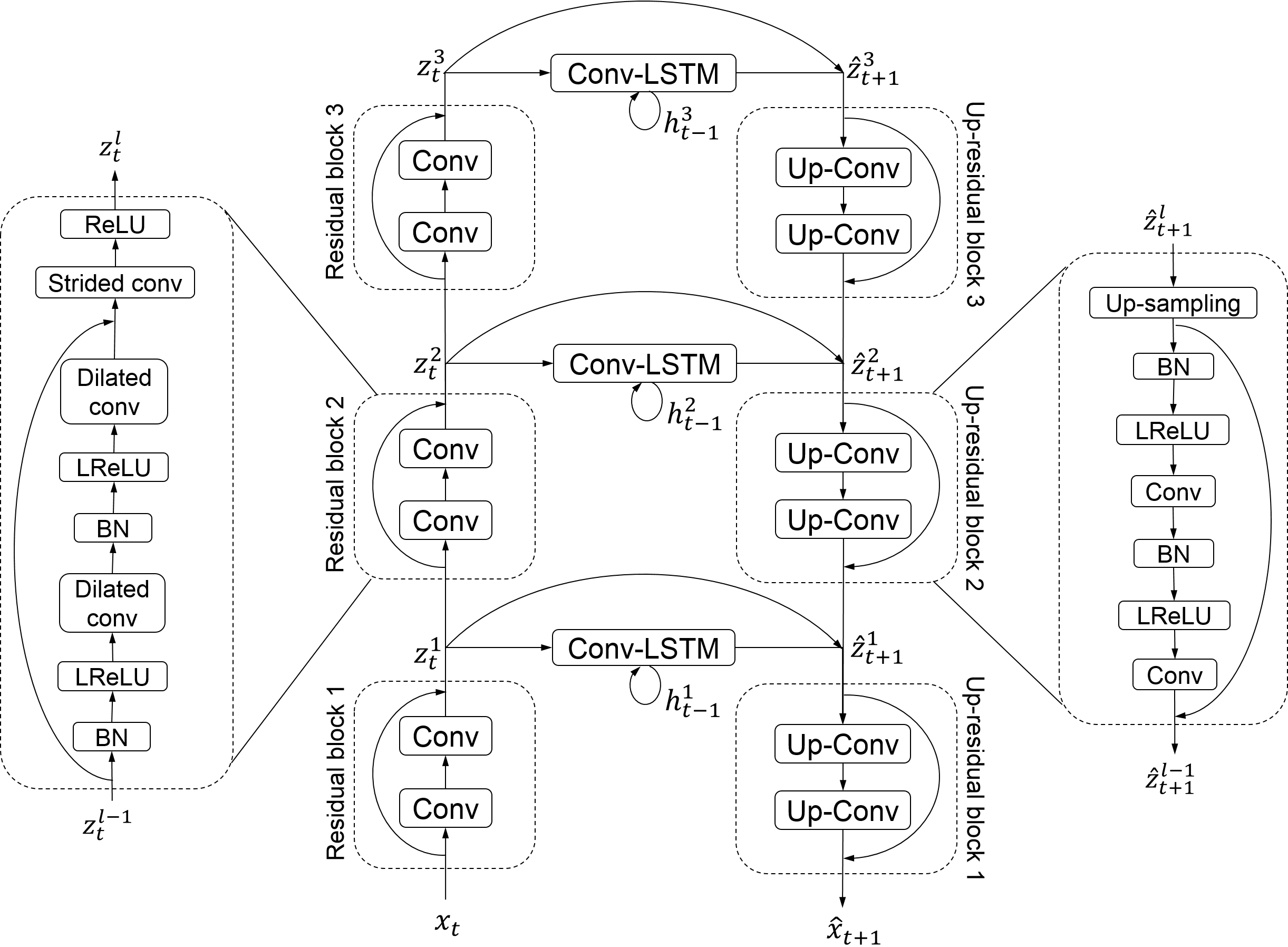}} 
  {\caption{Overview of VLN-ResNet, which incorporates ResNet-style residual blocks in the encoder and decoder.}\label{fig:overview_resnet}}  
\end{figure}

We have also experimented with a simple extension of the basic VLN model, by incorporating ResNet-style residual blocks (similar to those used in \citet{He2016}) into the encoder and decoder. This extended model, named VLN-ResNet, is shown in Figure \ref{fig:overview_resnet}.
VLN-ResNet has more layers than VLN but similar number of parameters. As can be seen in the figure, forward and backward signals can flow using multiple paths both in the encoder-decoder sides and in the lateral connections.

\subsection{Recurrent Residual Blocks}
\label{sec:connections}
\citet{valpola2015neural} showed that supplying the decoder with information from all encoder layers improves the learned image features and makes them more invariant to small changes in the input. VLN transfers this concept to unsupervised learning in the temporal domain, using both \textit{recurrent} lateral connections and \textit{feedforward} lateral connections. At each layer, a recurrent connection and a feedforward connection form a \textit{recurrent residual block}, where the recurrent connection represents the residual, and the feedforward connection represents the skip connection. Having a separate recurrent residual block at different layers in the hierarchy relieves the higher layers from the pressure of modeling lower-level spatial and temporal details. 
The feedforward lateral connections supply the decoder only with information about the latest input sample, and we expect them to be especially useful for modeling static parts.

Each recurrent connection consists of a convolutional Long Short-Term Memory (conv-LSTM) (\citet{Shi2015}), which is an extension of the fully-connected LSTM (\citet{hochreiter1997long}). In particular, conv-LSTMs replace the matrix multiplications with convolution operations and thus allow for preserving the spatial information.
At each time-step $t$ and for each encoder layer $l$, the input gate $i^l_t$, forget gate $f^l_t$, output gate $o^l_t$ and the candidate for the cell state $\tilde{c}^l_t$ of the convolutional-LSTM are computed as follows: 
\begin{align}
\label{eq:gates}
i^l_t = \sigma \left( z^l_t * W^l_{z i} + h^l_{t-1} * W^l_{h i} + b^l_i \right),\\
f^l_t = \sigma \left( z^l_t * W^l_{z f} + h^l_{t-1} * W^l_{h f} + b^l_f \right),\\
o^l_t = \sigma \left( z^l_t * W^l_{z o} + h^l_{t-1} * W^l_{h o} + b^l_o \right),\\
\tilde{c}^l_t = \tanh \left(z^l_t * W^l_{z\tilde{c}} +  h^l_{t-1} * W^l_{h \tilde{c}} + b^l_{\tilde{c}} \right),
\end{align}
where $z^l_t$ are the input feature maps from the $l$-th convolution layer, $h^l_{t-1}$ is the previous hidden state, $W^l_{z i}$, $W^l_{z f}$, $W^l_{z o}$, $W^l_{z\tilde{c}}$, $W^l_{h i}$, $W^l_{h f}$, $W^l_{h o}$, $W^l_{h \tilde{c}}$ are convolutional-kernel tensors, $b^l_i$, $b^l_f$, $b^l_o$, $b^l_{\tilde{c}}$ are bias terms, $\sigma$ and $\tanh$ are the sigmoid and hyperbolic tangent non-linearities, respectively, and $*$ is the convolution operator. Then, the hidden state $h^l_t$, output by a recurrent lateral connection at layer $l$, is computed as follows:
\begin{align}
\label{eq:cell_hidden}
h^l_t = o^l_t \odot \tanh \left( \tilde{c}^l_t \odot i^l_t + c^l_{t-1} \odot f^l_t \right),
\end{align}
where $\odot$ represents element-wise multiplication. 
The output of the recurrent connection $h^l_t$, of the feedforward connection $z^l_t$ and of the upper decoder layer $\tilde{z}^{l+1}_{t+1}$ are merged as follows:
\begin{align}
\label{eq:merging}
%\hat{z}^l_{t+1} = \text{LReLU} \left( \left( \text{LReLU} \left( \left( \tilde{z}^{l+1}_{t+1} %\oplus h^l_t \right) * W^l_h \right) \oplus z^l_t \right) * W^l_z \right),
\hat{z}^l_{t+1} = \text{LReLU} \left( \left( \text{LReLU} \left( \left( \tilde{z}^{l+1}_{t+1}, h^l_t \right) * W^l_h \right), z^l_t \right) * W^l_z \right),
\end{align}
where $\text{LReLU}$ is the leaky-ReLU non-linearity, $(\cdot, \cdot)$ denotes channel-wise concatenation, $W^l_h$ and $W^l_z$ are $(1,1)$ convolutional-kernel tensors. No batch-normalization is applied to the conv-LSTM.
 
\section{Experiments}
\label{sec:experiments}
\setcounter{footnote}{0}
We evaluate the Video Ladder Networks on the Moving MNIST dataset, that we generate using the code provided by the authors\footnote{\url{http://www.cs.toronto.edu/~nitish/unsupervised_video/}}. This dataset is derived from the popular MNIST dataset, which contains $60000$ training samples and $10000$ testing samples. Validation samples are selected by picking out $20\%$ training samples, while preserving the percentage of samples for each class. We produce the Moving MNIST samples from corresponding MNIST partitions on the fly, generating $10k$ train samples and $1k$ validation samples per epoch. Furthermore, evaluation is performed on the provided test-set of $10k$ samples.

Each video sequence is generated by moving two digits within the video frames. Each digit is randomly chosen and is initially positioned randomly inside a patch. It is then moved along a random direction with constant speed, bouncing off when hitting the frame's boundaries. Digits in the same frame move independently and are allowed to overlap. Each video sequence consists of $20$ frames, of which the first $10$ represent past frames and are used as input frames to the model, whereas the following $10$ frames represent the future ground-truth frames to be predicted. Each frame has size $64 \times 64$. 

\subsection{Model Architectures}
\label{subsec:architecture}
In VLN, the encoder consists of $3$ dilated convolutional layers with stride $(2,2)$, dilations $1,2,4$, and number of filters $32,64,96$. The decoder has $3$ convolutional layers preceded by up-sampling. Empirically, we observed that using dilation at the decoder does not bring any benefit. All layers use kernel size $(3,3)$. The convolutional LSTMs use same number of channels as in their input. In VLN-ResNet (shown in Figure \ref{fig:overview_resnet}), both the encoder and decoder consist of $3$ residual blocks, where each block contains two convolutional layers with no stride, kernel size $(3,3)$, and number of filters $28,28,58,58,90,90$. The encoder uses dilated convolutions, with dilation rates $1,2,2,4,4,8$, followed by element-wise sum of the skip connection and by a strided convolution for halving the resolution. At the decoder side, each residual block starts by up-sampling. We compare our models to most relevant prior art and to two baselines, VLN-BL and VLN-BL-FF. Specifically, VLN-BL has only one convolutional-LSTM at the top layer (with $128$ channels to match the number of parameters of VLN) and no feedforward lateral connections. VLN-BL-FF includes also feedforward connections at all layers.

In order to improve the computational efficiency of the model, we decided to extend the effective receptive field of the model by reducing the resolution via striding when going up in the model hierarchy. If computational efficiency is not an issue, it is possible to use larger dilations and/or more depth and preserve the resolution at all layers as in \citet{kalchbrenner2016video}. This is left as future work.

\subsection{Training and Evaluation}
\label{subsec:training}
We train and evaluate VLN using the \textit{binary cross-entropy} loss, by interpreting the grayscale targets as probabilities as in \citet{kalchbrenner2016video}. Ideally, the loss used for training should be averaged over $10$ predictions (as in prior works) but, due to limited computational resources, we average it over $5$ predictions. This is a disadvantage with respect to prior works. However, the losses used for model evaluation are averaged over $10$ frames, in order to fairly compare to the prior art. We train all our models for $1700$ epochs, using \textit{RMSprop} with initial learning rate $0.0001$ (except for VLN-ResNet, which is trained only for $1300$ epochs, with initial learning rate $0.0005$).

\begin{figure}[htbp]
  \centering
  {\includegraphics[width=0.6\linewidth]{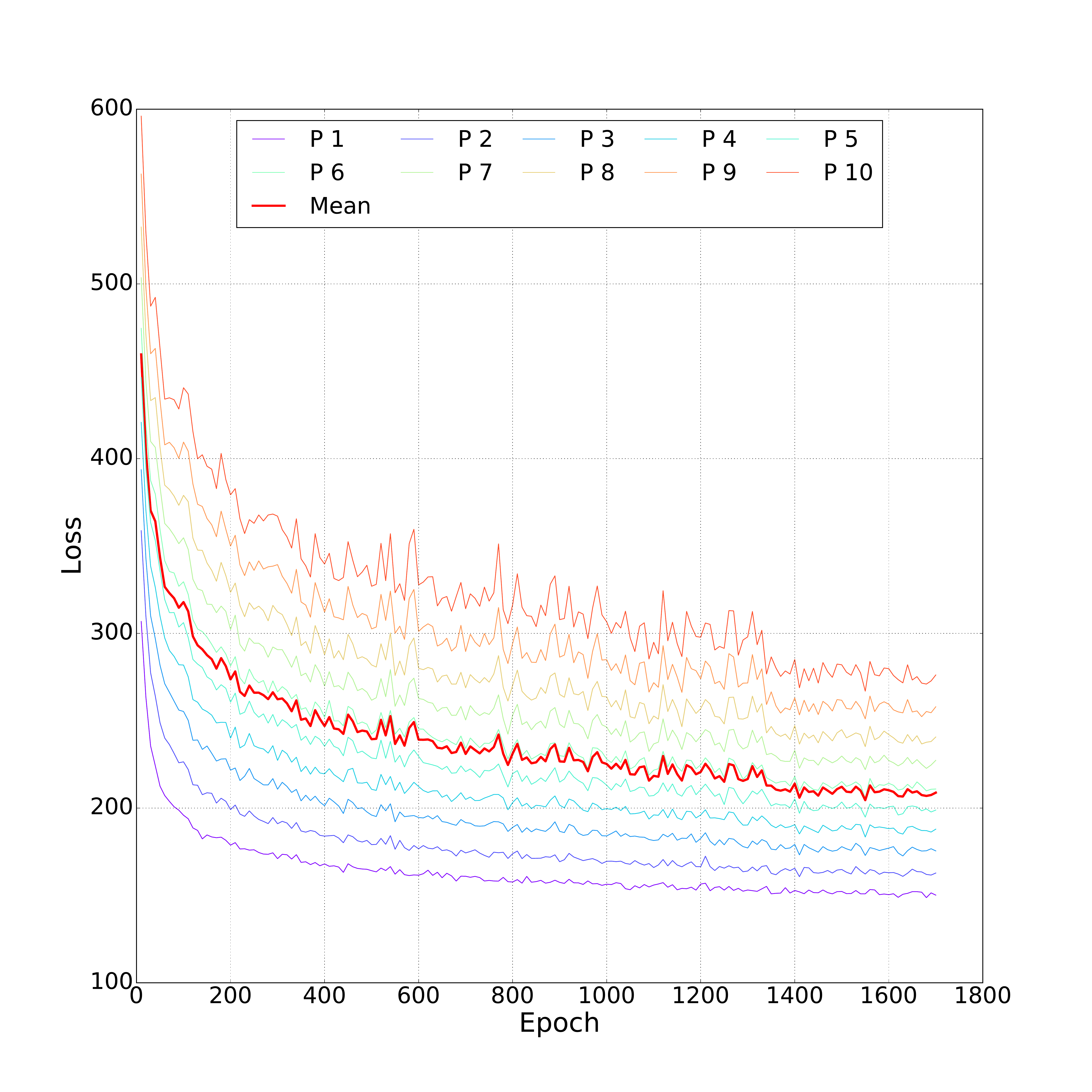}} 
  {\caption{Test-set loss for each time-step.}\label{fig:plot}}
\end{figure}

We use a windowing approach for training and prediction. The input window has length $10$. The latest prediction $\hat{x}_{t+1}$ is fed back to input and the window is shifted, so that the used ground-truth is only $x_{n_t+1:10}$ ($n_t$ is the number of fed back predictions at time $t$). When shifting the window, at training phase the conv-LSTM hidden state is preserved, whereas at evaluation phase the state is reset. See Figure \ref{fig:plot} for a plot of the test-set loss for each time-step. As can be seen, losses for the last $5$ predictions have more variance and are more far apart from each other, compared to the first $5$ predictions. This causes the average test loss to have high variance too.

\subsection{Results}
\label{subsec:results}
\begin{figure}[htbp]
  \centering
  {\includegraphics[width=1.0\linewidth]{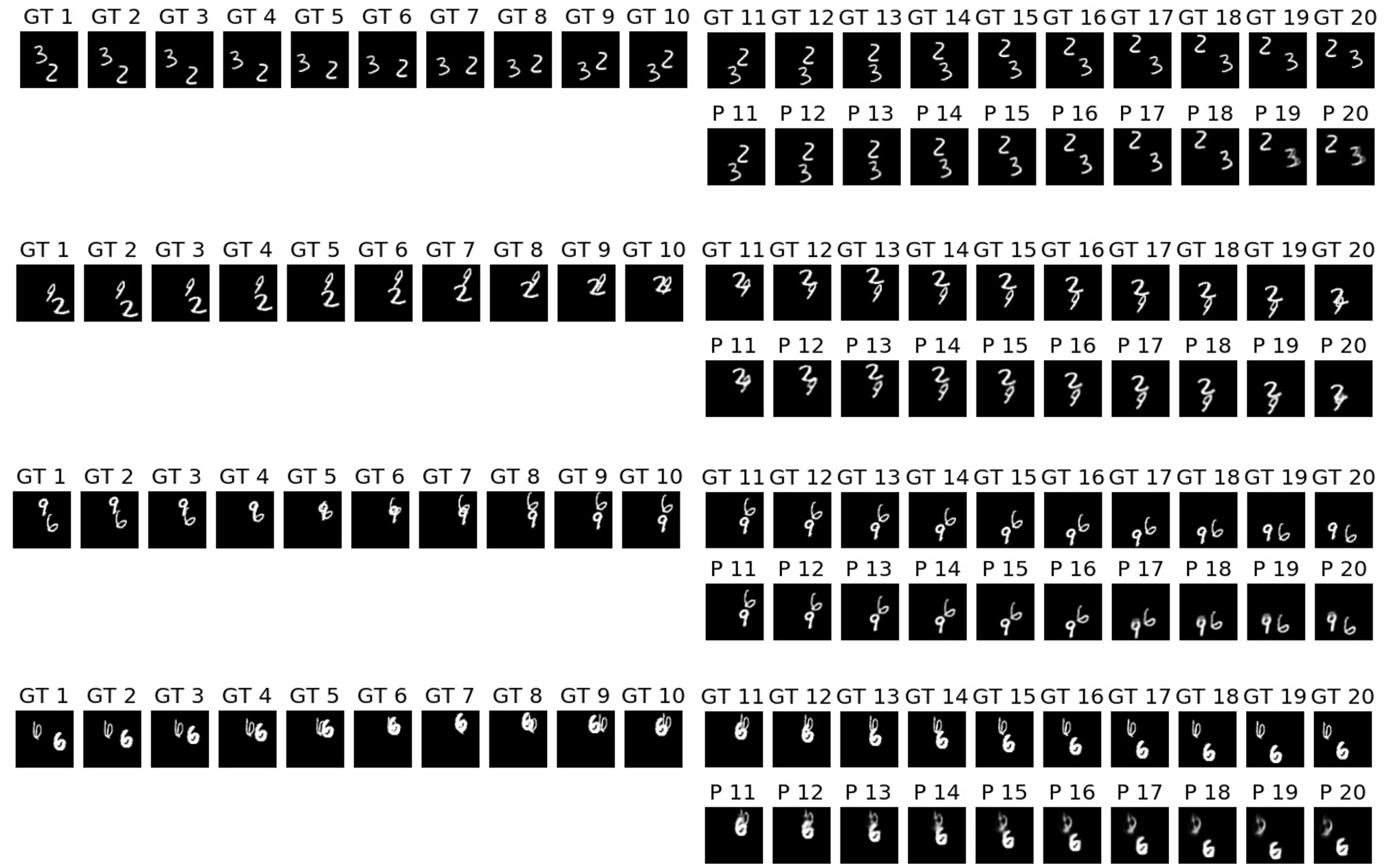}} 
  {\caption{Four randomly selected test-set samples. First row of each sample: the first $10$ frames are the input past frames and the remaining $10$ frames are the ground-truth future frames. Second row of each sample: future frames predicted by VLN-ResNet.}\label{fig:results}}
\end{figure}

The results in Table \ref{tab:results} show that VLN and VLN-ResNet outperform all other models except for the baseline of Video Pixel Networks (VPN-BL), which has much more parameters and depth. Our baselines, VLN-BL and VLN-BL-FF, perform well too and we attribute their competitive results to the use of convolutional-LSTM, batch-normalization, and our windowing training and prediction approach. By comparing VLN-BL and VLN-BL-FF, it is worth noticing that feedforward connections bring only a marginal improvement on the Moving MNIST dataset, providing evidence that the main benefit of VLN comes from the recurrent lateral connections. Figure \ref{fig:results} shows four randomly drawn test-set samples, where the predictions are generated by VLN-ResNet. 
\begin{table}
  \caption{Results on test-set (\mytilde { }means estimated from available information)}
  \label{tab:results}
  \centering
  \begin{tabular}{llll}
  \toprule
  \bfseries Model & \bfseries Test loss 				& \bfseries Depth & \bfseries \# Parameters\\
  \midrule
  conv-LSTM (\citet{Shi2015})								& 367.1			& Unknown  & 7.6M \\
  fcLSTM (\citet{srivastava2015unsupervised}) 				& 341.2					& \mytilde 2 & \mytilde 143M\\ 
  DFN (\citet{Brabandere2016nips})					& 285.2					& Unknown & 0.6M\\ 
  VPN-BL (\citet{kalchbrenner2016video})				& 110.1					& \mytilde 81 & \mytilde 30M\\
  \midrule  
  VLN-BL				& 222.3					& 7 & 1.2M\\
  VLN-BL-FF				& 220.1					& 8 & 1.2M\\
  VLN				& 207.0					& 9 & 1.2M\\  
  VLN-ResNet		& 187.7					& 15 & 1.3M\\  
  \bottomrule
  \end{tabular}
\end{table}

We left out from the comparison two other works which are not comparable to VLN. Specifically, in the VPN model described in \citet{kalchbrenner2016video}, each RGB frame is predicted pixel by pixel, which is very slow as it uses about $80$ layers and needs $N \times N \times 3$ iterations ($N \times N$ is the spatial resolution). The model described in \citet{Patraucean2016} encodes motion explicitly. Our model has simple structure, faster inference ($10$ iterations) and does not encode motion explicitly. Furthermore, VLN's recurrent residual blocks can be easily included in other encoder-decoder models, such as the VPN baseline. 

\section{Conclusions}
\label{sec:conclusions}
We proposed the Video Ladder Network, a novel neural architecture for generating future video frames, conditioned on past frames. VLN is characterized by its lateral recurrent residual blocks. These are lateral connections connecting the encoder and the decoder, and which summarize both spatial and temporal information at all layers. This way, the top layer is relieved from the pressure of modeling all spatial and temporal details, and the decoder can generate future frames by using information from different semantic levels. Furthermore, we proposed an extension of the VLN which uses ResNet-style encoder and decoder. The proposed models were evaluated on the Moving MNIST dataset, comparing them to prior works, including the current state-of-the-art model (\citet{kalchbrenner2016video}). We showed that VLN and VLN-ResNet achieve very competitive results, while having simple structure and providing fast inference.

\section*{References}

\bibliographystyle{plainnat}

\bibliography{vln_arxiv}

\end{document}